\documentclass[twocolumn]{fairmeta}

\usepackage[utf8]{inputenc}
\usepackage{amsmath,amssymb}
\usepackage{graphicx}
\usepackage{booktabs}
\usepackage{multirow}
\usepackage{placeins}
\usepackage{algorithm}
\usepackage{algpseudocode}
\usepackage{adjustbox}

\graphicspath{{figures/}}

\hypersetup{
  pdftitle={RoboForge: Cross-space Motion Refinement for Physically Executable Humanoid Whole-body Tracking},
  pdfauthor={Xichen Yuan, Zhe Li, Zhenzhe Zhang, Yangyang Wei, Xinying Guo, Shanghang Zhang, Jianfei Yang}
}

\title{RoboForge: Cross-space Motion Refinement for Physically Executable Humanoid Whole-body Tracking}

\author[1,*]{Xichen Yuan}
\author[1,*,\dagger]{Zhe Li}
\author[1,2,*]{Zhenzhe Zhang}
\author[1]{Yangyang Wei}
\author[1]{Xinying Guo}
\author[2]{Shanghang Zhang}
\author[1,\dagger]{Jianfei Yang}

\affiliation[1]{MARS Lab, Nanyang Technological University}
\affiliation[2]{Peking University}
\contribution[*]{Equal contribution}
\contribution[\dagger]{Corresponding authors}

\abstract{
Whole-body motion tracking has rapidly expanded the skill repertoire of humanoid robots, yet nearly all of this progress has come from stronger tracking policies, while the motion references they imitate are treated as fixed supervision. We argue that reference quality is a first-order factor in tracking performance. Human motion data carry capture and estimation artifacts, and retargeting adds physical inconsistencies due to the human--robot embodiment gap; discarding such motions sacrifices valuable skill content, whereas retaining them uncorrected undermines policy learning. We present RoboForge, a data-centric framework that resolves this dilemma by refining references instead of filtering them, repairing each error in the space where it arises. In human motion space, a quality-guided diffusion model localizes corrupted segments and inpaints only those segments, leaving the rest of the motion intact. In robot motion space, kinematic repair corrects foot-contact and self-collision defects, and annealed sampling-based optimization under physics rollouts turns kinematically plausible references into dynamically executable ones. RoboForge requires no change to the tracker: it improves three retargeting methods and four tracking policies, raising the success rate of BeyondMimic from 89.51\% to 96.20\% while reducing the mean per-joint position error (MPJPE-L) by 31.6\%. Refined references also accelerate policy learning, restore tracking performance from heavily corrupted inputs, and enable more than 30 high-dynamic motions, including flips and the Thomas flare, on a real Unitree G1.
}

\correspondence{\begin{tabular}[t]{@{}l@{}}Zhe Li (\email{LIZH0117@e.ntu.edu.sg}), Jianfei Yang (\email{jianfei.yang@ntu.edu.sg})\end{tabular}}

\begin{document}

\maketitle
\begin{figure}[t]
    \centering
    \includegraphics[
        width=\columnwidth,
        trim=86mm 26mm 148mm 18mm,
        clip
    ]{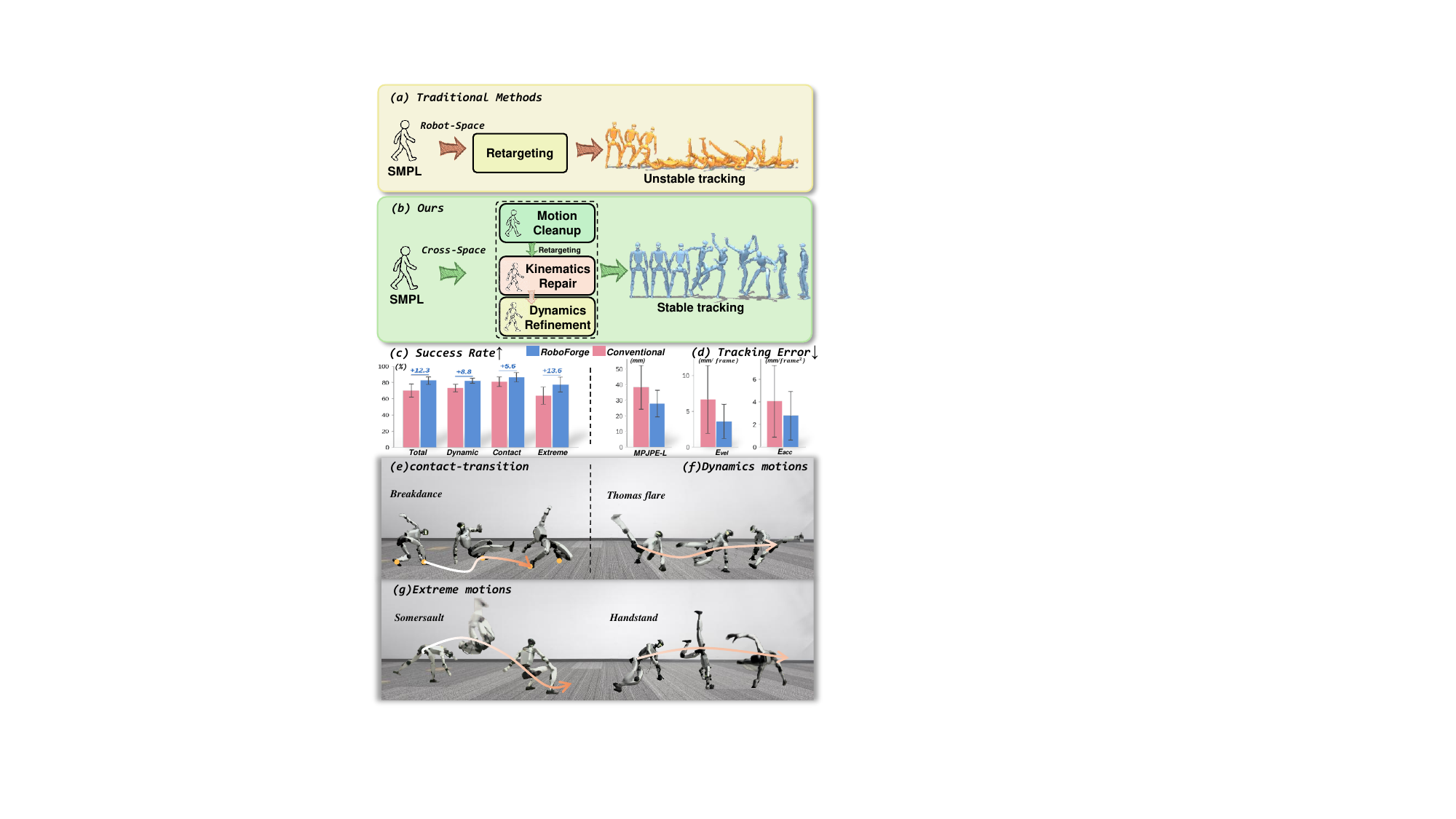}
    \caption{\textbf{Overview and evaluation of RoboForge.} Motion cleanup, kinematic repair, and dynamics refinement improve tracking success and reduce error relative to direct retargeting.}
    \label{fig:teaser}
\end{figure}
\section{Introduction}
\label{sec:intro}

Whole-body motion tracking has become the dominant route to expanding the skill repertoire of humanoid robots: human motions are retargeted into robot references, and a policy is trained to follow them~\citep{Human2Humanoid,GMT,li2025robomirror,li2025roboperform,li2025roboghost,li2026robodreamer}. Most recent progress has been policy-centric, with larger networks, better rewards, and more data. The reference itself, however, is almost always treated as ground truth. In practice it is not. Tracking performance depends not only on how capable the policy is, but on whether the reference it is asked to follow is physically achievable at all~\citep{GMR,li2026omega}.

Reference quality degrades at two distinct stages. At the source, motion capture and monocular pose estimation introduce jitter, foot skating, floating, and local temporal discontinuities~\citep{holden2018robust,shin2024wham}. During retargeting, the human--robot embodiment gap introduces kinematic inconsistencies, such as unstable foot contact and self-collision~\citep{GMR,OmniRetarget}, and dynamically infeasible trajectories that exceed the robot's actuation limits~\citep{SPIDER}, even when the source motion appears perfectly plausible. A stronger tracker cannot simply absorb these errors: it must follow the reference and remain balanced at the same time, and when the two goals conflict, infeasible references cause tracking failures and early episode termination during reinforcement learning (RL)~\citep{BeyondMimic,wang2026omnixtreme}, biasing the policy toward conservative execution at the expense of motion fidelity.

Existing remedies address this problem only partially. Optimized retargeting~\citep{zhao2026maketrackingeasy,SPIDER} and physics-guided reference refinement~\citep{omnitrack2026} operate exclusively on the robot side, so corruption inherited from the source motion is retargeted along with the motion rather than removed~\citep{StableMotion}. Scaling up motion libraries~\citep{humanoid-gpt26} increases coverage but not quality. Quality-based filtering~\citep{LIMMT} improves the average reference but discards exactly the motions that are hardest to capture cleanly, and cutting out locally corrupted segments breaks temporal coherence. This leaves a practical dilemma: discarding imperfect motions sacrifices valuable skill content, while retaining them uncorrected undermines policy learning.

The dilemma is sharpest for high-dynamic and extreme motions. They carry the richest information about contact patterns and whole-body coordination, and are therefore the most valuable for learning challenging skills~\citep{wang2026omnixtreme}; they are also the most likely to be corrupted at capture and the hardest to execute. With rapid contact transitions and large accelerations, a kinematically plausible reference can still produce large tracking errors or unstable contacts under proportional--derivative (PD) control, compounded by actuator limits and sim-to-real mismatch~\citep{he2025asap,wang2026omnixtreme}. Guaranteeing dynamic executability, not merely kinematic consistency, therefore requires evaluating references through PD-controlled physics rollouts, an optimization that is non-convex and, because of contact, non-smooth~\citep{SPIDER}. Sampling-based optimization suits this setting because it needs neither gradients nor convexity~\citep{mannor2003cross}, and annealing the sampling distribution moves the search from broad exploration to focused refinement~\citep{DIALMPC}.

Our central premise is that \emph{low-quality motions are not low-value motions}: imperfect references still contain useful skill content that can be recovered and made executable. We therefore introduce RoboForge, a data-centric framework that refines references across human and robot motion spaces, repairing each error in the space where it arises and is most directly observable (Fig.~\ref{fig:teaser}). In human motion space, a quality-guided diffusion model detects corrupted segments and inpaints only those segments, preserving the rest of the motion. After retargeting, robot-space refinement corrects foot-contact and self-collision defects, and then optimizes the PD control sequence through annealed sampling in a physics simulator, so that the final reference is the trajectory the robot can actually realize. The refined references are used to train existing tracking policies without any architectural change.

Our main contributions are summarized as follows:
\begin{itemize}
    \item \textbf{From filtering to refinement.}
    We treat motion quality as an optimization target rather than a selection criterion, recovering skill content from imperfect references instead of discarding it.

    \item \textbf{Error-aware cross-space refinement.}
    We repair source artifacts with quality-guided diffusion in human space, and embodiment-induced defects with kinematic repair and annealed physics-based optimization in robot space.

    \item \textbf{Consistent gains, tracker-agnostic.}
    RoboForge improves three retargeters and four tracking policies without modifying them, speeds up policy learning, restores performance under heavy corruption, and transfers to more than 30 motions on a real Unitree G1.
\end{itemize}

\section{Related Works}

\subsection{Human Motion Reconstruction and Refinement}

Recovering coherent human motion from noisy or incomplete observations is a long-standing problem, and the same concern applies to motions synthesized by language- or multimodal-conditioned generators~\citep{li2025lamp,li2026mulsmo}. HuMoR uses a learned motion prior for robust pose estimation~\citep{HuMoR}, and RoHM reconstructs motion under noise and occlusion with diffusion models~\citep{RoHM}. Closer to our setting, StableMotion learns to identify and repair corrupted segments from unpaired data~\citep{StableMotion}, and CleanMDM adds multimodal guidance for selective cleanup~\citep{CleanMDM}. Physics-aware approaches improve plausibility further: DiffPhy incorporates differentiable dynamics into reconstruction~\citep{DiffPhy}, and PhysDiff projects diffusion samples onto physically valid motion~\citep{PhysDiff}. All of these define plausibility with respect to the \emph{human} body, and a motion that is clean in SMPL space can still be unexecutable under the constraints of a specific robot. RoboForge uses human-space repair only as a first stage and follows it with robot-space physical refinement.

\subsection{Humanoid Retargeting and Physics-Based Refinement}

Retargeting transfers human motion to a humanoid embodiment under robot-specific kinematic constraints. GMR and RigUnification improve alignment across morphologies~\citep{GMR,RigUnification}, and contact-aware methods such as CoRe and OmniRetarget preserve contact and interaction consistency~\citep{CoRe,OmniRetarget}. Beyond kinematics, PHUMA combines data curation with physics-constrained retargeting~\citep{PHUMA}, SPARK enforces kinodynamic consistency through trajectory optimization~\citep{SPARK}, and SPIDER generates dynamically feasible trajectories via physics-guided sampling~\citep{SPIDER}. DIAL-MPC shows that annealed sampling balances global exploration and local refinement for nonlinear robot dynamics~\citep{DIALMPC}. All of these operate downstream of the source motion, so errors already present in the human data are inherited rather than corrected. RoboForge treats the source and the embodiment as two separate error sources and repairs both.

\subsection{Humanoid Motion Tracking}

Physics-based motion imitation enables controllers to acquire diverse skills from reference motions. DeepMimic established example-guided RL for tracking~\citep{DeepMimic}, AMP introduced adversarial motion priors~\citep{AMP}, PHC scaled imitation with failure recovery~\citep{PHC}, and MaskedMimic unified control through masked motion conditioning~\citep{MaskedMimic}. Recent humanoid systems bring these ideas to real hardware: GMT uses adaptive sampling and expert policies~\citep{GMT}, TWIST combines RL with behavior cloning~\citep{ze2025twist2}, and BeyondMimic enables agile and versatile real-world tracking~\citep{BeyondMimic}. All of these treat the reference as given. LIMMT is a notable exception, showing that quality-aware motion \emph{selection} yields better tracking with far less data~\citep{LIMMT}. RoboForge pushes this data-centric view one step further: rather than selecting the good references and discarding the rest, it repairs the imperfect ones, improving tracking without modifying the policy.

\begin{figure*}[t]
    \centering
    \includegraphics[
        width=\textwidth,
        trim=35mm 55mm 28mm 36mm,
        clip
    ]{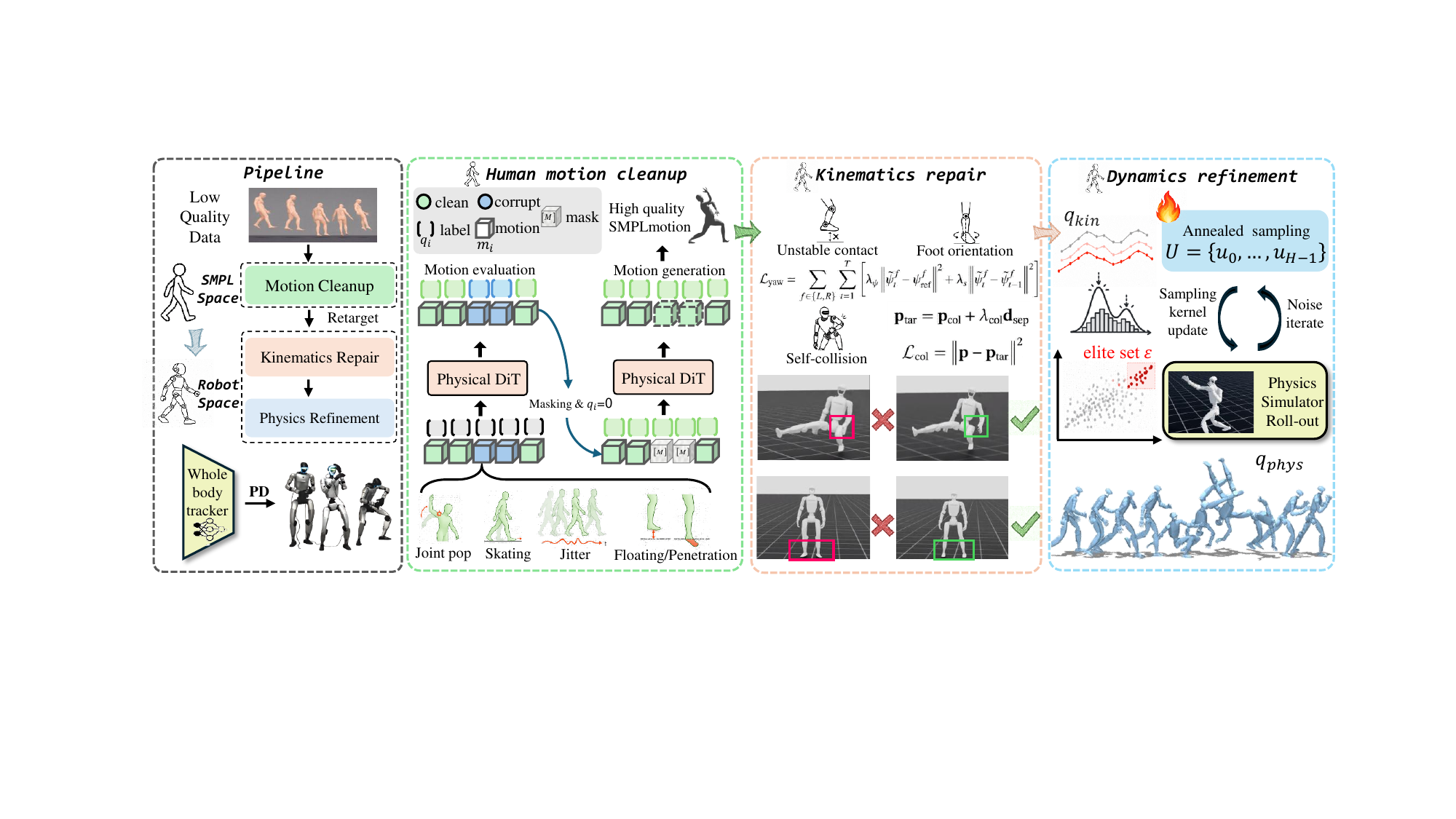}
    \caption{\textbf{The framework of RoboForge.}
    RoboForge refines motion references progressively across human and
    robot spaces. In human space, a shared Physical DiT first detects
    corrupted frames and then inpaints only those segments. After
    retargeting, kinematic repair corrects foot-contact, foot-orientation,
    and self-collision defects. Finally, annealed sampling optimizes the PD
    control sequence in a physics simulator, and the resulting rollout
    $q_{\mathrm{phys}}$ serves as the reference for downstream tracking.}
    \label{fig:framework}
\end{figure*}

\section{Method}

\subsection{Problem Formulation}

Reference errors arise in two spaces. \emph{Human-space} artifacts live in the SMPL motion itself and originate from capture, pose estimation, or generation. \emph{Robot-space} errors appear only after retargeting, because the robot differs from the human in morphology, joint limits, and actuation. No single stage can address both: robot-side optimization inherits source artifacts, while human-side cleanup is blind to the embodiment.

RoboForge therefore formulates reference refinement as a cross-space repair problem with one guiding principle: \emph{repair each error in the space where it arises and is most directly observable}. Given a raw SMPL motion, human-space repair produces a clean source motion; retargeting maps it to a nominal robot trajectory $q_{\mathrm{init}}$; kinematic repair corrects contact and geometric defects to obtain $q_{\mathrm{kin}}$; and dynamic refinement produces the physics-grounded reference $q_{\mathrm{phys}}$, which is used to train the tracking policy (Fig.~\ref{fig:framework}). Each stage has a distinct but complementary objective, and the pipeline is agnostic to the choice of retargeter and tracker.

\subsection{Human-Space Motion Repair}

Raw SMPL motions contain artifacts such as foot skating, floating or ground penetration, root and joint pops, and local temporal discontinuities. Two observations shape our design. First, these artifacts are usually confined to short segments, so regenerating the entire motion would modify frames that were never corrupted and risk altering the motion's content and semantics. Second, constraints specific to the robot, such as joint limits, actuation limits, and balance, are better handled later in robot space. Human-space repair is therefore \emph{local}: we regenerate only the corrupted segments and keep everything else unchanged.

We use a single diffusion transformer, which we call Physical DiT, trained with two inpainting modes that share the same model: one for motion-quality evaluation and one for motion generation. The predicted frame-wise quality determines which segments are regenerated.

\subsubsection{Artifact Detection}
Physical DiT is trained on unpaired SMPL motions with frame-wise quality labels, without any paired clean counterpart for corrupted samples. Each frame carries a binary quality indicator $q_i\in\{0,1\}$, with $q_i=0$ for clean and $q_i=1$ for corrupted. In the detection mode, the complete 232-D SMPL motion feature is kept and only the quality indicator is masked, so the model predicts motion quality directly from the motion sequence. It learns to recognize corrupted frames from root, joint, and foot motion patterns over time, without classifying artifact types explicitly. At inference, the predicted frame-wise score $\hat{q}_i$ is thresholded at $\tau=0.5$ to form a binary repair mask. Consecutive corrupted frames are merged into local repair regions, each extended by a small temporal margin on both sides to avoid boundary artifacts; all remaining frames are left untouched.

\subsubsection{Local Motion Inpainting}
In the generation mode, Physical DiT reconstructs masked motion features from the surrounding motion context and the frame-wise quality indicators. Training on mixed-quality unpaired data in this way teaches the model what clean motion looks like without paired supervision. At inference, only the detected segments are masked and their indicators are set to $q_i=0$, requesting clean motion. The surrounding clean frames serve as temporal context, so the regenerated segment is coherent with the original motion, and the structure and semantics of the sequence are preserved.

\subsection{Robot-Space Physical Refinement}

Even with a clean source, retargeting introduces robot-specific errors due to the embodiment gap: inconsistent foot contact, self-collision between links, and, for high-dynamic motions, trajectories that violate the robot's dynamics and actuator limits. We take the root pose and joint positions produced by any retargeter~\citep{GMR} as the nominal trajectory $q_{\mathrm{init}}$ and refine it in robot space, independently of the upstream retargeter, in two steps: kinematic repair yields $q_{\mathrm{kin}}$, and dynamic refinement keeps $q_{\mathrm{kin}}$ fixed while optimizing a realizable control sequence in simulation, whose rollout defines $q_{\mathrm{phys}}$.

\subsubsection{Kinematic Contact Repair}

We target two defects at this step: inconsistent foot contact and self-collision.

For foot contact, we first detect valid contact segments from the retargeted foot trajectories and partition the motion into stance and swing phases. We then perform gait optimization that removes stance-foot drift and foot--ground inconsistency during stance while leaving the swing-phase motion unchanged. We further stabilize the foot yaw during contact by optimizing
\begin{align}
\mathcal{L}_{\mathrm{yaw}}
={}&
\sum_{f,t}\lambda_{\psi} c_t^f
\left|\tilde{\psi}_t^f-\psi_{\mathrm{ref}}^f\right|^2
+\sum_{f,t}\lambda_s
\left|\tilde{\psi}_t^f-\tilde{\psi}_{t-1}^f\right|^2
\notag\\
&+
\sum_{f,t}\lambda_{\alpha}
\left[
\max\left(0,
|\ddot{\tilde{\psi}}_t^f|-\alpha_{\max}
\right)
\right]^2,
\label{eq:foot_yaw_loss}
\end{align}
where $\tilde{\psi}_{t}^{f}$ is the optimized yaw of foot $f$ at frame $t$, $\psi_{\mathrm{ref}}^{f}$ is the reference yaw of the corresponding contact segment, and $c_t^f\in\{0,1\}$ indicates contact. The first term keeps the foot yaw consistent during contact, the second suppresses abrupt frame-to-frame rotation, and the third penalizes yaw angular acceleration above $\alpha_{\max}$; $\lambda_{\psi}$, $\lambda_{s}$, and $\lambda_{\alpha}$ weight the three terms.

For self-collision, we detect interpenetrating links and push each colliding body part along its separation direction. The target position is
\begin{equation}
\mathbf{p}_{\mathrm{tar}}
=
\mathbf{p}_{\mathrm{col}}
+
\lambda_{\mathrm{col}}
\mathbf{d}_{\mathrm{sep}},
\label{eq:collision_target}
\end{equation}
and the collision-repair loss is
\begin{equation}
\mathcal{L}_{\mathrm{col}}
=
\left\|
\mathbf{p}
-
\mathbf{p}_{\mathrm{tar}}
\right\|^{2},
\label{eq:collision_loss}
\end{equation}
where $\mathbf{p}_{\mathrm{col}}$ is the current position of the colliding part, $\mathbf{d}_{\mathrm{sep}}$ the separation direction, $\lambda_{\mathrm{col}}$ the offset magnitude, and $\mathbf{p}$ the corrected position. The joint configuration is adjusted locally within the robot's joint limits so as to resolve the collision with minimal deviation from the original motion, and the corrected trajectory is smoothed to avoid abrupt changes between frames. The result, $q_{\mathrm{kin}}$, is free of the main contact and geometric defects; whether it can be executed under the robot's dynamics is the subject of the next stage.

\subsubsection{Physics-Sampling Dynamic Refinement}
\label{sec:dynamic_refinement}

\begin{algorithm}[!t]
\caption{Annealed Control Sampling}
\label{alg:annealed_sampling}
\small
\begin{algorithmic}[1]
\Require Initial control sequence $U^0$, number of samples $N$,
number of iterations $K$
\State $\mu^{(0)} \gets U^0$
\For{$k=0,\ldots,K-1$}
    \State Sample $N$ candidates $U_n^{(k)}$ around $\mu^{(k)}$ via Eq.~\eqref{eq:candidate_control}
    \For{$n=1,\ldots,N$}
        \State Roll out $U_n^{(k)}$ in simulation and evaluate $R_n$
    \EndFor
    \State Select high-reward candidates as the elite set $\mathcal{E}$
    \If{$k<K-1$}
        \State Update $\mu^{(k+1)}$ by reward-weighted averaging over $\mathcal{E}$
        \State Reduce the perturbation scale $\sigma_{k+1}<\sigma_k$
    \Else
        \State $U^\ast \gets \arg\max_n R_n$
    \EndIf
\EndFor
\State \Return $U^\ast$
\end{algorithmic}
\end{algorithm}

Kinematic feasibility does not imply dynamic executability. A trajectory can satisfy every geometric constraint and still be unrealizable because inertia, contact, and actuator limits govern what the robot can do under PD control, which is precisely the failure mode of high-dynamic and extreme motions. We therefore keep $q_{\mathrm{kin}}$ fixed as the target and optimize the joint-position control sequence $U$ that, when executed in a physics simulator, produces a rollout as close as possible to it. Here $u_t$ denotes the PD position target and $q_t$ the simulated robot state.

Over a rolling horizon of $H$ control steps, we define
\begin{equation}
U=\{u_0,\ldots,u_{H-1}\}.
\label{eq:control_sequence}
\end{equation}
At iteration $k$, the current sequence $\mu^{(k)}$ serves as the center of the sampling kernel. Gaussian perturbations are drawn at sparse control nodes and interpolated over the full horizon, yielding $N$ candidate sequences
\begin{equation}
U_n^{(k)}
=
\mu^{(k)}
+
\epsilon_n^{(k)},
\qquad
\epsilon_n^{(k)}
\sim
\mathcal{N}\!\left(0,\sigma_k^2 I\right).
\label{eq:candidate_control}
\end{equation}
All candidates are rolled out from the same simulator state and scored against the fixed $q_{\mathrm{kin}}$ with the cost
\begin{equation}
\begin{aligned}
J(U)
=
\sum_{t=0}^{H-1}
\big[\,
&w_q e_q(t)
+
w_v e_v(t)\\
&+
w_c e_c(t)
+
w_u e_u(t)
\,\big],
\end{aligned}
\label{eq:physics_sampling_cost}
\end{equation}
where $e_q$ and $e_v$ measure pose and velocity tracking error, $e_c$ penalizes contact inconsistency with respect to the contact schedule established during kinematic repair, and $e_u$ regularizes the control sequence against excessive deviation and actuator effort. The sampling reward is $R(U)=-J(U)$.

After each round of rollouts, the high-reward candidates form an elite set that updates the kernel center by reward-weighted averaging, and the perturbation scale $\sigma_k$ is reduced according to an annealing schedule, so that the search shifts from broad exploration to local refinement (Algorithm~\ref{alg:annealed_sampling}). At the final iteration, the highest-reward candidate is committed, the horizon advances from the simulated state reached by the committed controls, and the procedure repeats until the full control sequence $U^\ast$ is obtained. We finally replay $U^\ast$ from the initial state and define the resulting state trajectory as the physics-grounded reference $q_{\mathrm{phys}}$. By construction, $q_{\mathrm{phys}}$ is a motion the robot model has already executed, so the tracker is never asked to follow something the robot cannot do.

\FloatBarrier
\section{Experiments}

Our experiments build the case for data-centric refinement step by step, around four questions:

\textbf{Q1:} Does RoboForge improve downstream tracking without changing the retargeter or the tracker?

\textbf{Q2:} Where do the gains come from, i.e., what does each stage contribute?

\textbf{Q3:} Are refined references easier to learn from?

\textbf{Q4:} How far does it go: can it recover references from heavily corrupted inputs, and do the gains transfer to a real robot?

\subsection{Experimental Setup}
\noindent
\textit{1) Motion Libraries:}
We build motion libraries from AMASS~\citep{AMASS:ICCV:2019} and MotionDecode~\citep{MotionDecode}, comprising approximately 4.5K motion episodes with diverse styles. A fixed held-out library is used for all simulation-based evaluations, and corrupted counterparts of the same motions are constructed for the robustness study. For real-robot deployment on a Unitree G1, we further curate 30 representative motions covering high-dynamic movements, frequent contact transitions, and extreme whole-body motions.

\noindent
\textit{2) Metrics:}
Following \citet{SONIC,OmniH2O}, we report success rate (SR), the percentage of motions completed without triggering the failure condition; MPJPE-L (mm), the mean 3D position error of the evaluated body joints between executed and reference motions; and the corresponding joint velocity and acceleration errors $E_{\mathrm{vel}}$ (mm/frame) and $E_{\mathrm{acc}}$ (mm/frame$^2$). Together they capture execution success, motion fidelity, and temporal consistency.

\subsection{Motion Tracking Performance (Q1)}
\label{sec:motion_tracking}

To test whether better references help regardless of the surrounding pipeline, we run two complementary comparisons that disentangle retargeting from tracking: we apply RoboForge to different retargeters with a fixed tracker, and to a fixed retargeter with different frozen trackers.

\begin{table}[t]
\centering
\caption{Success rates (\%) across motion categories and retargeters, with and without RoboForge.}
\label{tab:retargeter_comparison}
\vspace{-1mm}
\scriptsize
\setlength{\tabcolsep}{3.8pt}
\renewcommand{\arraystretch}{1.08}
\resizebox{\columnwidth}{!}{
\begin{tabular}{lccccc}
\toprule
\textbf{Retargeter}
& \textbf{RoboForge}
& \textbf{Total}
& \textbf{Dynamic}
& \textbf{Contact}
& \textbf{Extreme} \\
\midrule
\multirow{2}{*}{PHC~\citep{PHC}}
& $\times$ & 62.1 & 67.9 & 74.5 & 52.6 \\
& $\checkmark$ & \textbf{77.4} & \textbf{79.6} & \textbf{80.2} & \textbf{67.5} \\
\midrule
\multirow{2}{*}{PhySINK~\citep{PHUMA}}
& $\times$ & 78.4 & 77.2 & 85.7 & 73.6 \\
& $\checkmark$ & \textbf{86.2} & \textbf{81.1} & \textbf{91.2} & \textbf{79.4} \\
\midrule
\multirow{2}{*}{GMR~\citep{GMR}}
& $\times$ & 70.2 & 74.6 & 82.8 & 65.4 \\
& $\checkmark$ & \textbf{83.9} & \textbf{85.4} & \textbf{88.3} & \textbf{85.6} \\
\bottomrule
\end{tabular}
}
\end{table}

\begin{table*}[!t]
\centering
\caption{Tracking performance of four frozen trackers on raw and RoboForge-refined references (GMR retargeting).}
\label{tab:tracker_generalization}
\small
\setlength{\tabcolsep}{6pt}
\renewcommand{\arraystretch}{1.08}
\begin{adjustbox}{max width=\textwidth}
\begin{tabular}{lccccc}
\toprule
\textbf{Tracker} &
\textbf{RoboForge} &
\textbf{SR $\uparrow$} &
\textbf{MPJPE-L $\downarrow$} &
\textbf{$E_{\mathrm{vel}} \downarrow$} &
\textbf{$E_{\mathrm{acc}} \downarrow$} \\
& & (\%) & (mm) & (mm/frame) & (mm/frame$^2$) \\
\midrule
\multirow{2}{*}{TWIST2~\citep{ze2025twist2}}
& $\times$ & 79.32$\pm$0.65 & 58.647$\pm$2.245 & 13.576$\pm$0.042 & 8.682$\pm$0.004 \\
& $\checkmark$ & \textbf{87.43$\pm$0.32} & \textbf{39.334$\pm$1.132} & \textbf{7.025$\pm$0.081} & \textbf{5.864$\pm$0.002} \\
\midrule
\multirow{2}{*}{RoboPerform~\citep{li2025roboperform}}
& $\times$ & 86.11$\pm$2.42 & 35.312$\pm$8.564 & 5.733$\pm$0.036 & 3.324$\pm$0.005 \\
& $\checkmark$ & \textbf{94.43$\pm$3.35} & \textbf{28.354$\pm$3.548} & \textbf{3.249$\pm$0.061} & \textbf{2.336$\pm$0.002} \\
\midrule
\multirow{2}{*}{GMT~\citep{GMT}}
& $\times$ & 84.26$\pm$5.13 & 30.547$\pm$6.683 & 4.013$\pm$0.054 & 2.312$\pm$0.006 \\
& $\checkmark$ & \textbf{95.26$\pm$3.32} & \textbf{24.436$\pm$7.228} & \textbf{2.422$\pm$0.025} & \textbf{1.947$\pm$0.003} \\
\midrule
\multirow{2}{*}{BeyondMimic~\citep{BeyondMimic}}
& $\times$ & 89.51$\pm$1.81 & 28.258$\pm$2.247 & 3.266$\pm$0.075 & 1.818$\pm$0.008 \\
& $\checkmark$ & \textbf{96.20$\pm$1.23} & \textbf{19.332$\pm$0.852} & \textbf{1.683$\pm$0.038} & \textbf{0.945$\pm$0.004} \\
\bottomrule
\end{tabular}
\end{adjustbox}
\end{table*}

\paragraph{Across Retargeters}
Table~\ref{tab:retargeter_comparison} applies RoboForge on top of three upstream retargeters, PHC~\citep{PHC}, PhySINK~\citep{PHUMA}, and GMR~\citep{GMR}, replays the resulting references with the same tracker, and reports success rates on three stress categories: high-dynamic motions, which challenge dynamic feasibility; contact-transition motions, which stress contact consistency; and extreme whole-body motions, which amplify embodiment-induced kinematic infeasibility. RoboForge raises the overall success rate of every retargeter, from 62.1\% to 77.4\% for PHC, 78.4\% to 86.2\% for PhySINK, and 70.2\% to 83.9\% for GMR, and improves every category for every retargeter. Two patterns stand out. The largest absolute gains occur where the baseline is weakest, PHC (+15.3 points overall) and extreme motions under GMR (+20.2 points), which is exactly where reference infeasibility, rather than tracker capacity, limits performance. And because the three categories fail for different reasons, consistent gains across all of them indicate that the three refinement stages are complementary rather than redundant.

\paragraph{Across Trackers}
Table~\ref{tab:tracker_generalization} fixes GMR as the retargeter and evaluates four frozen trackers, TWIST2~\citep{ze2025twist2}, RoboPerform~\citep{li2025roboperform}, GMT~\citep{GMT}, and BeyondMimic~\citep{BeyondMimic}, on the same motion set and protocol. RoboForge improves every tracker on every metric. The weakest tracker benefits most, with TWIST2 reducing MPJPE-L from 58.6 to 39.3~mm and halving $E_{\mathrm{vel}}$, yet the strongest tracker, BeyondMimic, still gains 6.69 points in SR and 31.6\% in MPJPE-L. The average SR gain across the four policies is 8.53 points. Reference quality therefore acts as a ceiling that no tracker, however capable, can exceed on its own, and raising that ceiling benefits all of them.

\subsection{Ablation Studies (Q2)}
\label{sec:ablation}

Having established that RoboForge helps, we ask where the gains come from. We enable the three stages progressively, human-space repair, kinematic repair, and annealed dynamic refinement, with the same retargeter and tracker throughout~\citep{GMR,BeyondMimic}.

\begin{table*}[t]
\centering
\caption{Stage-wise ablation of RoboForge. Each row adds one stage to the previous row.}
\label{tab:component_ablation}
\small
\setlength{\tabcolsep}{8pt}
\renewcommand{\arraystretch}{1.15}
\begin{adjustbox}{max width=\textwidth}
\begin{tabular}{lcccc}
\toprule
\textbf{Method}
& \shortstack{\textbf{SR}$\uparrow$\\(\%)}
& \shortstack{\textbf{MPJPE-L}$\downarrow$\\(mm)}
& \shortstack{$\mathbf{E_{vel}}\downarrow$\\(mm/frame)}
& \shortstack{$\mathbf{E_{acc}}\downarrow$\\(mm/frame$^2$)} \\
\midrule
Retargeted
& $89.51\pm1.81$ & $28.258\pm2.247$ & $3.266\pm0.075$ & $1.818\pm0.008$ \\
+ Human-Space Repair
& $90.14\pm2.12$ & $23.442\pm1.354$ & $2.236\pm0.041$ & $1.372\pm0.001$ \\
+ Kinematic Repair
& $93.75\pm0.58$ & $20.518\pm1.016$ & $1.946\pm0.027$ & $1.027\pm0.002$ \\
\textbf{+ Annealed Dynamics (Full)}
& $\mathbf{96.20\pm1.23}$ & $\mathbf{19.332\pm0.852}$ & $\mathbf{1.683\pm0.038}$ & $\mathbf{0.945\pm0.004}$ \\
\bottomrule
\end{tabular}
\end{adjustbox}
\end{table*}

\begin{figure}[!t]
    \centering
    \includegraphics[
        width=\columnwidth,
        trim=290bp 45bp 330bp 45bp,
        clip
    ]{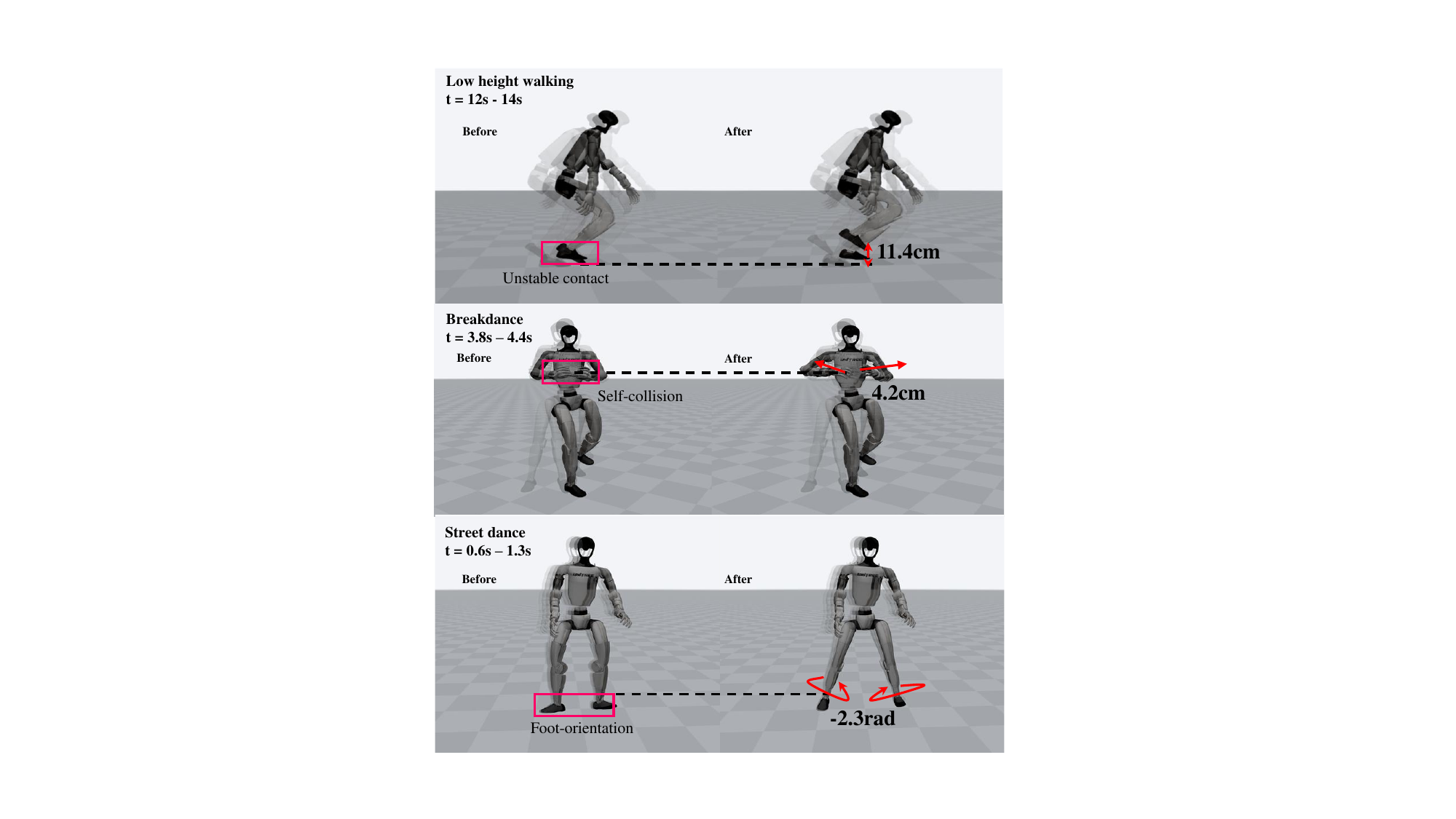}
    \caption{\textbf{Kinematic repair.} Before and after correction of unstable foot contact (low-height walking), self-collision (breakdance), and foot orientation (street dance).}
    \label{fig:kinematic_repair}
\end{figure}

\textbf{Stage-wise contribution.}
Table~\ref{tab:component_ablation} shows that the three stages act on different failure modes. Human-space repair reduces MPJPE-L sharply, from 28.258 to 23.442~mm, and cuts $E_{\mathrm{vel}}$ by a third, but barely moves SR (89.51\% to 90.14\%): cleaning the source improves the fidelity of motions the robot can already execute, yet does not by itself make infeasible motions feasible. Kinematic repair delivers the first large jump in SR, to 93.75\%, confirming that contact and self-collision defects introduced by retargeting are a primary cause of outright failure. Dynamic refinement adds a further 2.45 points, reaching 96.20\%, and yields the lowest error on every metric. The full pipeline improves SR by 6.69 points over raw retargeting and reduces MPJPE-L, $E_{\mathrm{vel}}$, and $E_{\mathrm{acc}}$ by 31.6\%, 48.5\%, and 48.0\%, respectively. Each stage is necessary: none of the three alone accounts for the full gain.

\textbf{Kinematic repair.}
Fig.~\ref{fig:kinematic_repair} shows representative corrections of robot-specific defects: foot--ground consistency is restored in low-height walking, self-collision is resolved in breakdance, and abnormal foot orientation is corrected in street dance. In each case the correction is local and the overall motion structure is retained, consistent with the targeted nature of the repair.

\begin{figure}[!t]
    \centering
    \includegraphics[
        width=0.98\columnwidth,
        trim=182bp 3bp 259bp 5bp,
        clip
    ]{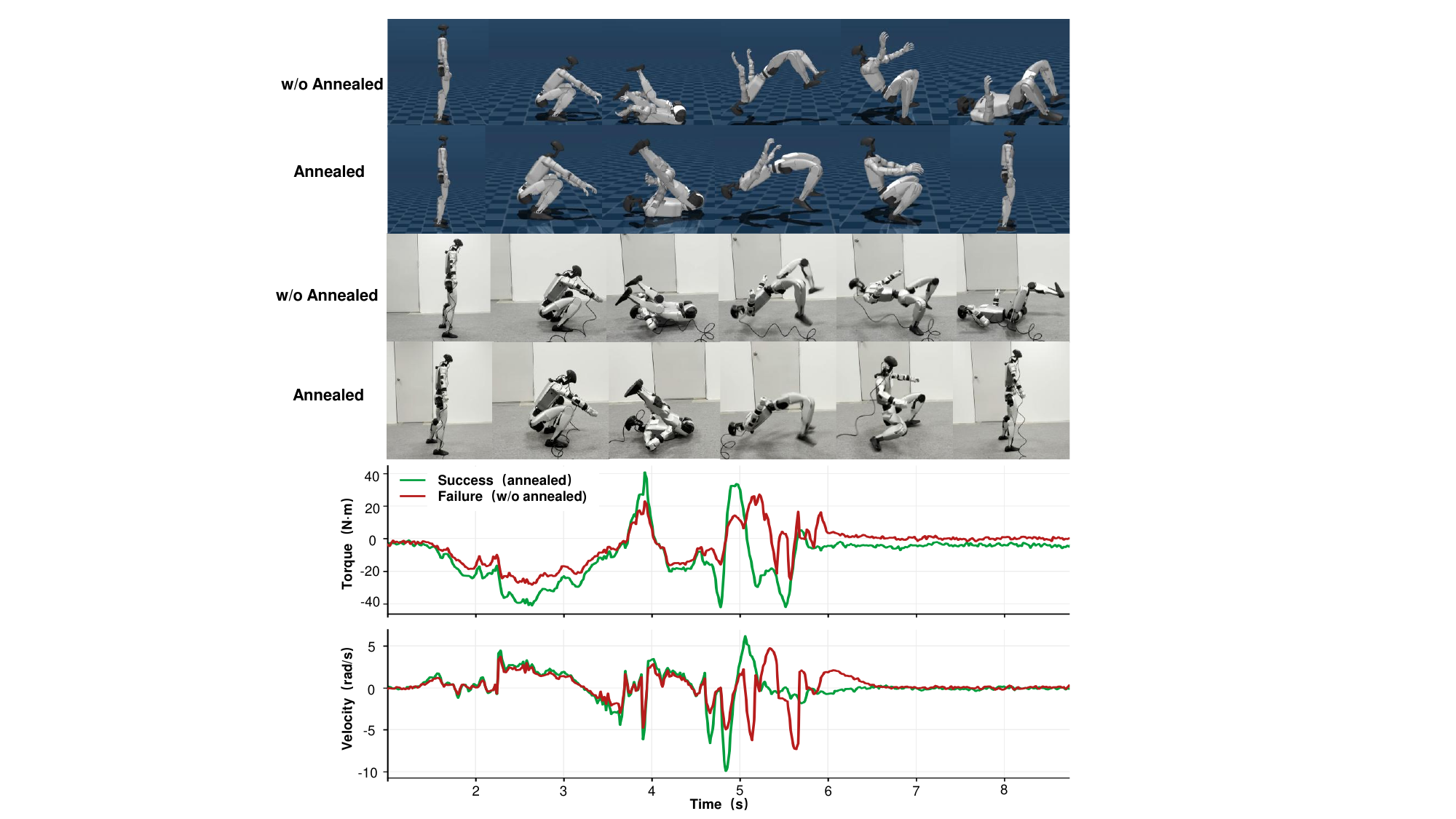}
    \caption{\textbf{Effect of annealed sampling on a high-dynamic motion.} Execution with and without annealing in simulation and on the real robot, with the corresponding waist joint torque and angular-velocity profiles.}
    \label{fig:waist_motor}
    \vspace{-5mm}
\end{figure}

\textbf{Annealed dynamic refinement.}
Fig.~\ref{fig:waist_motor} isolates the role of annealing on a challenging high-dynamic motion. Sampling-based optimization without annealing fails to complete the motion, whereas annealed sampling executes it and recovers to a stable standing posture, both in simulation and on the real robot. The waist actuator profiles explain why: with annealing, torque and angular velocity rise to support the critical dynamic transition and then settle; without it, the failed execution is followed by pronounced torque and velocity oscillations. Annealing is what allows the search to find a control sequence that is realizable within actuator limits, which is the capability that kinematic correction alone cannot provide.

\subsection{Reference Learnability (Q3)}
\label{sec:reference_learnability}

\begin{figure}[t]
    \centering
    \includegraphics[
        width=0.64\columnwidth,
        trim=71mm 42mm 90mm 27mm,
        clip
    ]{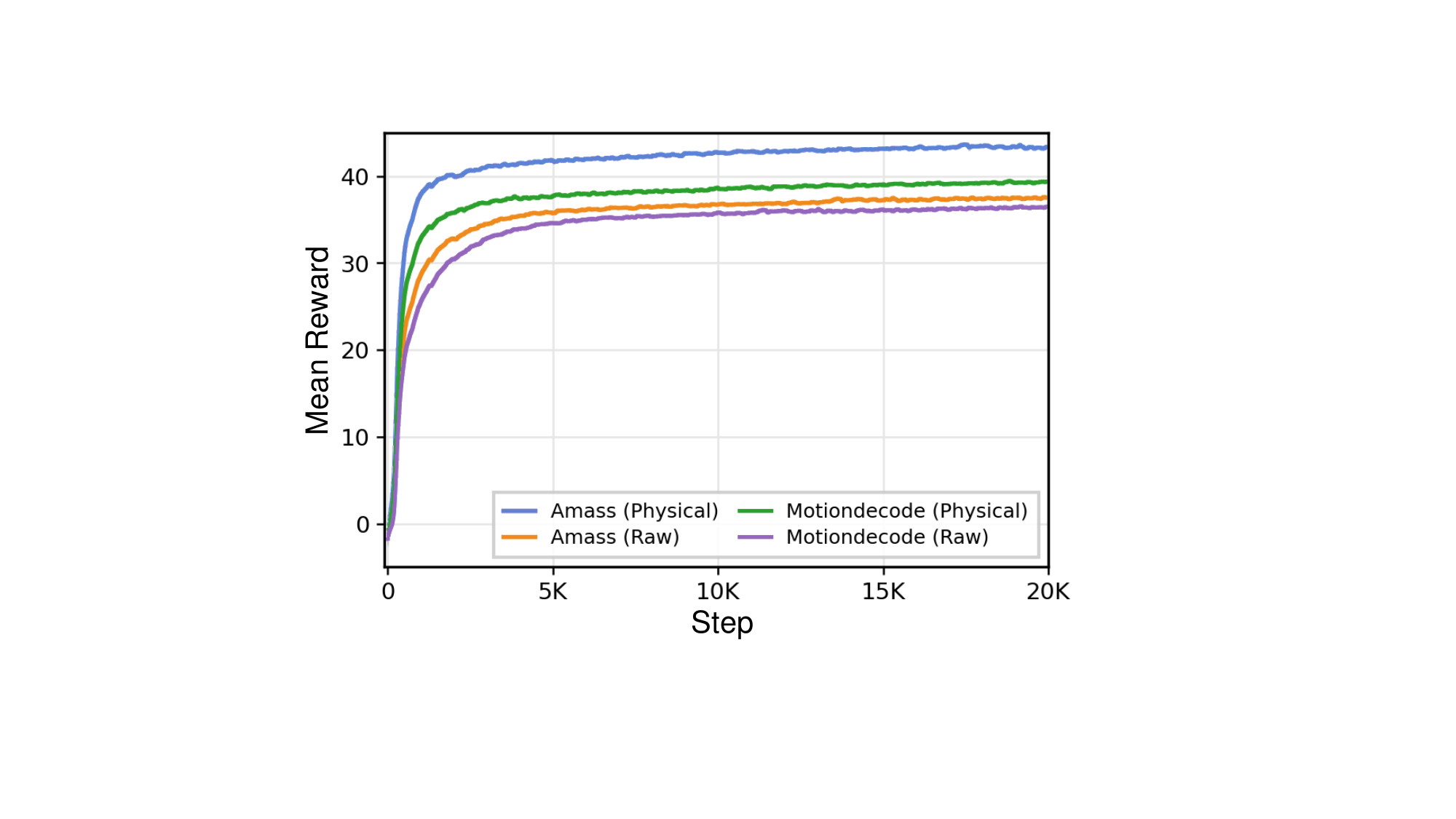}
    \caption{Learning curves of the same tracking policy trained on raw and RoboForge-refined references from AMASS and MotionDecode~\citep{AMASS:ICCV:2019,MotionDecode}.}
    \label{fig:mean_reward}
\end{figure}

The results so far evaluate frozen trackers. A stronger claim is that refined references are also better \emph{training} data. We train the same policy~\citep{BeyondMimic} from scratch on raw retargeted references and on their refined counterparts, on both AMASS~\citep{AMASS:ICCV:2019} and MotionDecode~\citep{MotionDecode}. As Fig.~\ref{fig:mean_reward} shows, refined references lead to faster convergence and a higher final reward on both datasets. This matches the mechanism described in Sec.~\ref{sec:intro}: infeasible references cause early termination and push the policy toward conservative behavior, whereas executable references provide a consistent learning signal. The gain appears on two datasets with different capture setups, so it stems from the refinement itself rather than from a particular dataset.

\subsection{Robustness to Corrupted Inputs (Q4)}
\label{sec:corrupted_motion}

\begin{table*}[!t]
\centering
\caption{Robustness to corrupted motion inputs. Robot-space metrics (left) and SMPL-space metrics (right). For Jitter, values closer to the clean ground truth are better.}
\label{tab:corrupted_motion}
\vspace{-1mm}
\scriptsize
\setlength{\tabcolsep}{3.3pt}
\renewcommand{\arraystretch}{1.08}
\resizebox{\textwidth}{!}{
\begin{tabular}{lcccc|cccc}
\toprule
\multirow{2}{*}{\textbf{Input / Processing}}
& \multicolumn{4}{c}{\textbf{Robot Space}}
& \multicolumn{4}{c}{\textbf{SMPL Space}} \\
\cmidrule(lr){2-5}
\cmidrule(lr){6-9}
& \shortstack{SR $\uparrow$}
& \shortstack{MPJPE-L $\downarrow$}
& \shortstack{{$E_{\mathrm{vel}}\downarrow$}}
& \shortstack{{$E_{\mathrm{acc}}\downarrow$}}
& {FS Dist $\downarrow$}
& {Jitter $\rightarrow$}
& \shortstack{{GMPJPE $\downarrow$} }
& {M2M Score $\uparrow$} \\
\midrule
Clean GT (AMASS)
& 89.51$\pm$1.81
& 28.258$\pm$2.247
& 3.266$\pm$0.075
& 1.818$\pm$0.008
& 2.62
& 1.88
& 0.000
& 1.000 \\
Corrupted Input (BrokenAMASS)
& 62.46$\pm$3.57
& 46.463$\pm$1.649
& 7.384$\pm$0.061
& 3.632$\pm$0.011
& 5.47
& 8.24
& 18.024
& 0.843 \\
RoboForge
& \textbf{94.32$\pm$2.04}
& \textbf{25.651$\pm$1.807}
& \textbf{3.303$\pm$0.062}
& \textbf{1.436$\pm$0.038}
& \textbf{3.12$\pm$0.04}
& \textbf{1.79$\pm$0.06}
& \textbf{4.32$\pm$0.005}
& \textbf{0.914$\pm$0.008} \\
\bottomrule
\end{tabular}
}
\vspace{-1mm}
\end{table*}

\begin{figure}[!t]
    \centering
    \includegraphics[
        width=0.98\columnwidth,
        height=0.42\textheight,
        keepaspectratio,
        trim=100mm 12mm 100mm 7mm,
        clip
    ]{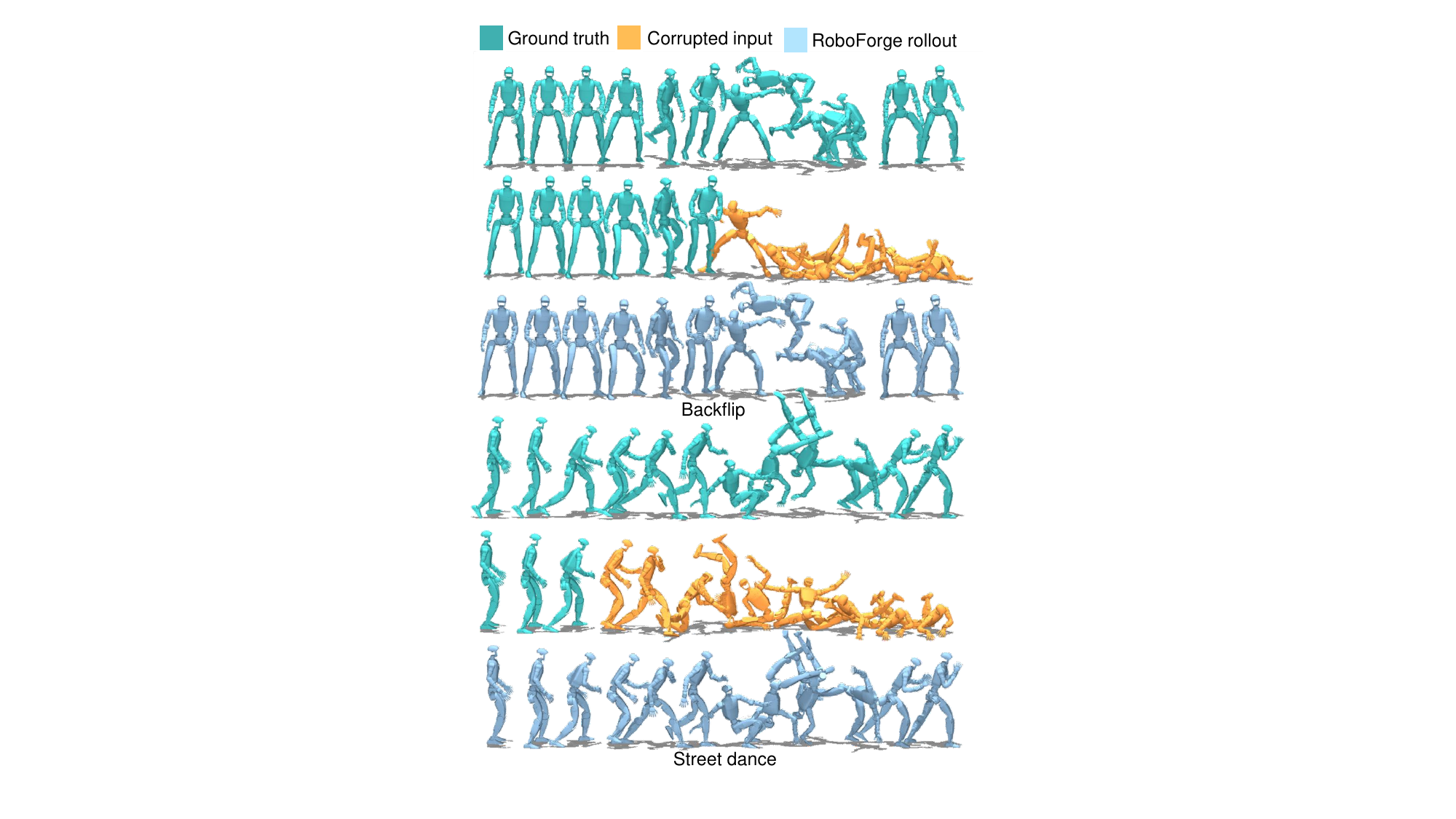}
    \caption{Recovery from corrupted inputs. Ground-truth motion, corrupted input, and RoboForge-refined rollout on two dynamic motions, backflip and street dance.}
    \label{fig:brokenrobot}
    \vspace{-5mm}
\end{figure}

The experiments above start from ordinary motion data. We now ask how much corruption RoboForge can absorb. Following the BrokenAMASS protocol of StableMotion~\citep{StableMotion}, we corrupt AMASS motions with four artifact types, jittering, foot sliding, over-smoothing, and drifting, and pass clean and corrupted motions through the same retargeting and tracking pipeline. We evaluate in both spaces: in robot space with SR, MPJPE-L, $E_{\mathrm{vel}}$, and $E_{\mathrm{acc}}$ as before, and in SMPL space with the four metrics of \citet{StableMotion}, namely FS Dist (foot sliding), Jitter (the derivative of acceleration), GMPJPE (global joint-position error against the paired clean ground truth), and M2M Score (cosine similarity to the ground truth in the latent space of a pretrained TMR model, which captures semantic preservation).

Table~\ref{tab:corrupted_motion} tells a two-part story. First, corruption at the source propagates all the way to execution: SR drops from 89.51\% to 62.46\% and MPJPE-L rises from 28.258 to 46.463~mm, confirming that source-motion quality, not only retargeting, bounds tracking performance. Second, RoboForge reverses the damage. It restores SR to 94.32\% and MPJPE-L to 25.651~mm, reduces FS Dist from 5.47 to 3.12 and Jitter from 8.24 to 1.79, lowers GMPJPE from 18.02 to 4.32~cm, and raises the M2M Score from 0.843 to 0.914. The improvement in GMPJPE and M2M Score is the important one: it shows that the corrupted content is \emph{recovered}, not merely smoothed away. Notably, RoboForge from corrupted input matches, and on several metrics exceeds, direct retargeting of clean AMASS, because it repairs source corruption and robot-space infeasibility at the same time. Fig.~\ref{fig:brokenrobot} illustrates this recovery on a backflip and a street dance.

\subsection{Real-World Experiments (Q4)}
\label{sec:real_world}
\begin{figure}[!t]
    \centering
    \includegraphics[
        width=\columnwidth,
        trim=265 145 350 135,
        clip
    ]{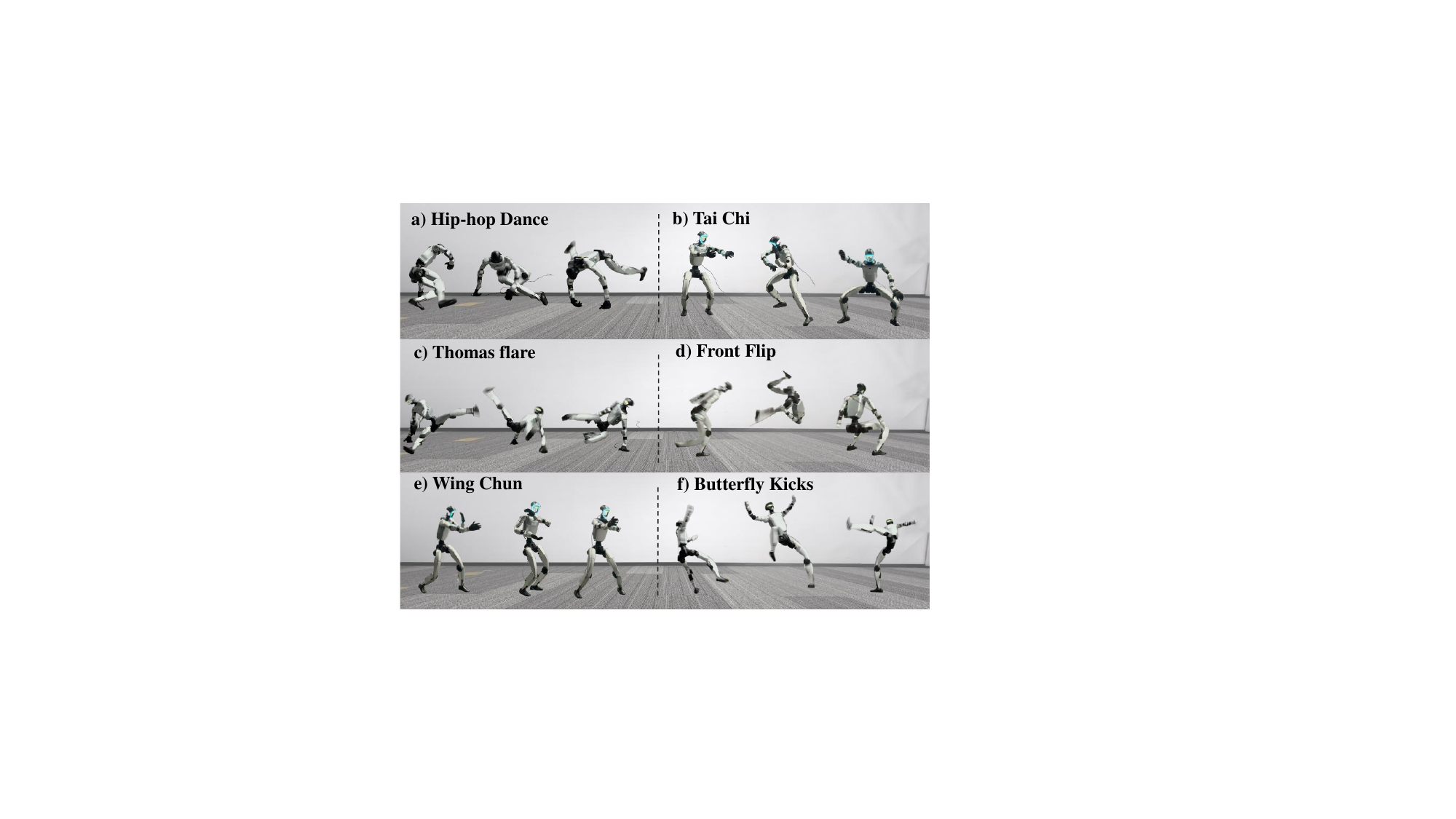}
    \caption{Real-world results on a Unitree G1. Representative examples from more than 30 deployed motions, including hip-hop dance, Tai Chi, Thomas flare, and front flip.}
    \label{fig:real_robot}
\end{figure}

Finally, we deploy policies trained on RoboForge-refined references on a real Unitree G1 using motions with diverse dynamics and contact patterns. More than 30 motions are executed successfully; Fig.~\ref{fig:real_robot} shows representative examples including hip-hop dance, Tai Chi, the Thomas flare, and a front flip, and Fig.~\ref{fig:waist_motor} shows the annealing comparison on hardware. The gains in reference executability observed in simulation therefore carry over to physical hardware across a broad range of whole-body motions.

\section{Conclusions}
We presented RoboForge, a data-centric framework built on a simple premise: imperfect motion references should be repaired, not discarded. By combining quality-guided diffusion in human motion space with kinematic repair and annealed physics-based optimization in robot motion space, it corrects each error where it arises and delivers references the robot can actually execute. Without modifying any tracker, RoboForge improves three retargeting methods and four tracking policies (average success-rate gain of 8.53 points), accelerates policy learning on AMASS and MotionDecode, recovers clean-level tracking from heavily corrupted inputs, and transfers to more than 30 motions on a real Unitree G1. These results position motion quality as an optimization target that compounds with advances in tracking policies, rather than a filtering criterion. Future work will explore preference-guided quality assessment and tracking-aware feedback to enable scalable, adaptive physics-based refinement of large-scale humanoid motion datasets.

\bibliographystyle{assets/plainnat}
\bibliography{mybib}

\begin{thebibliography}{42}
\providecommand{\natexlab}[1]{#1}
\providecommand{\url}[1]{\texttt{#1}}
\expandafter\ifx\csname urlstyle\endcsname\relax
  \providecommand{\doi}[1]{doi: #1}\else
  \providecommand{\doi}{doi: \begingroup \urlstyle{rm}\Url}\fi

\bibitem[Araujo et~al.(2025)Araujo, Ze, Xu, Wu, and Liu]{GMR}
Joao~Pedro Araujo, Yanjie Ze, Pei Xu, Jiajun Wu, and C.~Karen Liu.
\newblock Retargeting matters: General motion retargeting for humanoid motion
  tracking.
\newblock \emph{arXiv preprint arXiv:2510.02252}, 2025.

\bibitem[Chen et~al.(2025)Chen, Ji, Cheng, Peng, Peng, and Wang]{GMT}
Zixuan Chen, Mazeyu Ji, Xuxin Cheng, Xuanbin Peng, Xue~Bin Peng, and Xiaolong
  Wang.
\newblock {GMT}: General motion tracking for humanoid whole-body control.
\newblock \emph{arXiv preprint arXiv:2506.14770}, 2025.

\bibitem[{CHINGMU}(2026)]{MotionDecode}
{CHINGMU}.
\newblock {ChingMu} 1000-hour embodied motion dataset.
\newblock \url{https://huggingface.co/datasets/CMRobot/MotionDecode}, 2026.
\newblock MotionDecode dataset. Accessed: Oct. 20, 2026.

\bibitem[G{\"a}rtner et~al.(2022)G{\"a}rtner, Andriluka, Coumans, and
  Sminchisescu]{DiffPhy}
Erik G{\"a}rtner, Mykhaylo Andriluka, Erwin Coumans, and Cristian Sminchisescu.
\newblock Differentiable dynamics for articulated {3D} human motion
  reconstruction.
\newblock In \emph{CVPR}, 2022.

\bibitem[Guan et~al.(2026)Guan, Qi, Lin, Chen, Zhang, Wang, Yu, Wang, and
  Yi]{LIMMT}
Yu~Guan, Zekun Qi, Chenghuai Lin, Xuchuan Chen, Wenyao Zhang, Jilong Wang,
  Xinqiang Yu, He~Wang, and Li~Yi.
\newblock {LIMMT}: Less is more for motion tracking.
\newblock In \emph{Proceedings of the 43rd International Conference on Machine
  Learning}, volume 306 of \emph{Proceedings of Machine Learning Research},
  pages 37504--37517. PMLR, 2026.

\bibitem[He et~al.(2025{\natexlab{a}})He, Gao, Xiao, Zhang, Wang, Wang, Luo,
  He, Sobanbab, Pan, et~al.]{he2025asap}
Tairan He, Jiawei Gao, Wenli Xiao, Yuanhang Zhang, Zi~Wang, Jiashun Wang,
  Zhengyi Luo, Guanqi He, Nikhil Sobanbab, Chaoyi Pan, et~al.
\newblock {ASAP}: Aligning simulation and real-world physics for learning agile
  humanoid whole-body skills.
\newblock \emph{arXiv preprint arXiv:2502.01143}, 2025{\natexlab{a}}.

\bibitem[He et~al.(2025{\natexlab{b}})He, Luo, He, Xiao, Zhang, Zhang, Kitani,
  Liu, and Shi]{OmniH2O}
Tairan He, Zhengyi Luo, Xialin He, Wenli Xiao, Chong Zhang, Weinan Zhang,
  Kris~M. Kitani, Changliu Liu, and Guanya Shi.
\newblock {OmniH2O}: Universal and dexterous human-to-humanoid whole-body
  teleoperation and learning.
\newblock In \emph{Proceedings of the Conference on Robot Learning}, volume 270
  of \emph{Proceedings of Machine Learning Research}, pages 1516--1540. PMLR,
  2025{\natexlab{b}}.

\bibitem[Holden(2018)]{holden2018robust}
Daniel Holden.
\newblock Robust solving of optical motion capture data by denoising.
\newblock \emph{ACM Transactions on Graphics}, 37\penalty0 (4):\penalty0
  165:1--165:12, 2018.
\newblock \doi{10.1145/3197517.3201302}.

\bibitem[Huang et~al.(2026)Huang, Yuan, Gu, Fang, Zhang, Wang, Gao, and
  Zhang]{Human2Humanoid}
Tianchen Huang, Feiyang Yuan, Junchi Gu, Shurui Fang, Xiaohu Zhang, Yu~Wang,
  Wei Gao, and Shiwu Zhang.
\newblock {Human2Humanoid}: Physics-aware cross-morphology motion retargeting
  for humanoid robots.
\newblock \emph{arXiv preprint arXiv:2606.03476}, 2026.

\bibitem[Jeong et~al.(2025{\natexlab{a}})Jeong, Byun, Kim, Choi, Oh, Lee,
  Darwish, Kim, and Choi]{RigUnification}
Taemoon Jeong, Taehyun Byun, Jihoon Kim, Keunjun Choi, Jaesung Oh, Sungpyo Lee,
  Omar Darwish, Joohyung Kim, and Sungjoon Choi.
\newblock Robust and expressive humanoid motion retargeting via
  optimization-based rig unification.
\newblock In \emph{IEEE/RSJ International Conference on Intelligent Robots and
  Systems}, pages 21619--21626, 2025{\natexlab{a}}.
\newblock \doi{10.1109/IROS60139.2025.11246607}.

\bibitem[Jeong et~al.(2025{\natexlab{b}})Jeong, Chai, Choi, Bak, Kim, Yoon,
  Lee, Lee, Lee, Kim, and Choi]{CoRe}
Taemoon Jeong, Yoonbyung Chai, Sol Choi, Jaewan Bak, Chanwoo Kim, Jihwan Yoon,
  Yisoo Lee, Jongwon Lee, Kyungjae Lee, Joohyung Kim, and Sungjoon Choi.
\newblock {CoRe}: A hybrid approach of contact-aware optimization and learning
  for humanoid robot motions.
\newblock In \emph{2025 IEEE-RAS 24th International Conference on Humanoid
  Robots (Humanoids)}, pages 293--300, 2025{\natexlab{b}}.
\newblock \doi{10.1109/Humanoids65713.2025.11203055}.

\bibitem[Lee et~al.(2025)Lee, Kim, Park, Kim, Hwang, Lee, and Choo]{PHUMA}
Kyungmin Lee, Sibeen Kim, Minho Park, Hyunseung Kim, Dongyoon Hwang, Hojoon
  Lee, and Jaegul Choo.
\newblock {PHUMA}: Physically-grounded humanoid locomotion dataset.
\newblock \emph{arXiv preprint arXiv:2510.26236}, 2025.

\bibitem[Li et~al.(2026{\natexlab{a}})Li, Zhi, Wang, Liu, Yan, Liu, Wang, Jia,
  and Huang]{omnitrack2026}
Yuhan Li, Peiyuan Zhi, Yunshen Wang, Tengyu Liu, Sixu Yan, Wenyu Liu, Xinggang
  Wang, Baoxiong Jia, and Siyuan Huang.
\newblock {OmniTrack}: General motion tracking via physics-consistent
  reference.
\newblock \emph{arXiv preprint arXiv:2602.23832}, 2026{\natexlab{a}}.

\bibitem[Li et~al.(2025{\natexlab{a}})Li, Chi, Wei, Zhu, Huang, Sun, Peng,
  Wang, Wang, Liu, Xu, and Zhang]{li2025roboperform}
Zhe Li, Cheng Chi, Yangyang Wei, Boan Zhu, Tao Huang, Zhenguo Sun, Yibo Peng,
  Pengwei Wang, Zhongyuan Wang, Fangzhou Liu, Chang Xu, and Shanghang Zhang.
\newblock Do you have freestyle? {E}xpressive humanoid locomotion via audio
  control.
\newblock \emph{arXiv preprint arXiv:2512.23650}, 2025{\natexlab{a}}.

\bibitem[Li et~al.(2025{\natexlab{b}})Li, Chi, Wei, Zhu, Peng, Huang, Wang,
  Wang, Zhang, and Xu]{li2025roboghost}
Zhe Li, Cheng Chi, Yangyang Wei, Boan Zhu, Yibo Peng, Tao Huang, Pengwei Wang,
  Zhongyuan Wang, Shanghang Zhang, and Chang Xu.
\newblock From language to locomotion: Retargeting-free humanoid control via
  motion latent guidance.
\newblock \emph{arXiv preprint arXiv:2510.14952}, 2025{\natexlab{b}}.

\bibitem[Li et~al.(2025{\natexlab{c}})Li, Chi, Zhu, Wei, Bai, Ji, Peng, Huang,
  Wang, Wang, et~al.]{li2025robomirror}
Zhe Li, Cheng Chi, Boan Zhu, Yangyang Wei, Shuanghao Bai, Yuheng Ji, Yibo Peng,
  Tao Huang, Pengwei Wang, Zhongyuan Wang, et~al.
\newblock {RoboMirror}: Understand before you imitate for video to humanoid
  locomotion.
\newblock \emph{arXiv preprint arXiv:2512.23649}, 2025{\natexlab{c}}.

\bibitem[Li et~al.(2025{\natexlab{d}})Li, Yuan, He, Qiu, Zhu, Gu, Shen, Dong,
  Dong, and Yang]{li2025lamp}
Zhe Li, Weihao Yuan, Yisheng He, Lingteng Qiu, Shenhao Zhu, Xiaodong Gu,
  Weichao Shen, Yuan Dong, Zilong Dong, and Laurence Yang.
\newblock {LaMP}: Language-motion pretraining for motion generation, retrieval,
  and captioning.
\newblock In \emph{International Conference on Learning Representations},
  volume 2025, pages 84238--84250, 2025{\natexlab{d}}.

\bibitem[Li et~al.(2026{\natexlab{b}})Li, He, Zhong, Shen, Zuo, Qiu, Zhu, Dong,
  Yang, Xu, et~al.]{li2026mulsmo}
Zhe Li, Yisheng He, Lei Zhong, Weichao Shen, Qi~Zuo, Lingteng Qiu, Shenhao Zhu,
  Zilong Dong, Laurence~T Yang, Chang Xu, et~al.
\newblock {MulSMo}: Multimodal stylized motion generation by bidirectional
  control flow.
\newblock \emph{IEEE Transactions on Image Processing}, 2026{\natexlab{b}}.

\bibitem[Li et~al.(2026{\natexlab{c}})Li, Wang, Cai, and Fu]{CleanMDM}
Zhe Li, Shicheng Wang, Bowen Cai, and Huan Fu.
\newblock {CleanMDM}: Clean motion diffusion model for multimodal motion
  cleanup.
\newblock \emph{arXiv preprint arXiv:2610.05411}, 2026{\natexlab{c}}.

\bibitem[Li et~al.(2026{\natexlab{d}})Li, Wei, Yuan, Zhang, Yuan, Zhang, and
  Yang]{li2026robodreamer}
Zhe Li, Yangyang Wei, Xichen Yuan, Zhenzhe Zhang, Weihao Yuan, Shanghang Zhang,
  and Jianfei Yang.
\newblock {RoboDreamer}: Anticipatory humanoid locomotion with predictive
  state-space models.
\newblock \emph{arXiv preprint arXiv:2609.07096}, 2026{\natexlab{d}}.

\bibitem[Li et~al.(2026{\natexlab{e}})Li, Zhang, Wei, Zhang, Yuan, Zhi, Li,
  Guo, Gao, Yang, et~al.]{li2026omega}
Zhe Li, Zhenzhe Zhang, Yangyang Wei, Wenjie Zhang, Xichen Yuan, Peiyuan Zhi,
  Gen Li, Xinying Guo, Fengjie Gao, Jianfei Yang, et~al.
\newblock {$\omega$-0: A Latent Predictive World Action Model for Concurrent
  Humanoid Loco-Manipulation}.
\newblock \emph{arXiv preprint arXiv:2608.06375}, 2026{\natexlab{e}}.

\bibitem[Liao et~al.(2026)Liao, Truong, Huang, Gao, Tevet, Sreenath, and
  Liu]{BeyondMimic}
Qiayuan Liao, Takara~E. Truong, Xiaoyu Huang, Yuman Gao, Guy Tevet, Koushil
  Sreenath, and C.~Karen Liu.
\newblock {BeyondMimic}: From motion tracking to versatile humanoid control via
  guided diffusion.
\newblock \emph{Science Robotics}, 11\penalty0 (117):\penalty0 eadx8924, 2026.
\newblock \doi{10.1126/scirobotics.adx8924}.

\bibitem[Luo et~al.(2023)Luo, Cao, Winkler, Kitani, and Xu]{PHC}
Zhengyi Luo, Jinkun Cao, Alexander Winkler, Kris Kitani, and Weipeng Xu.
\newblock Perpetual humanoid control for real-time simulated avatars.
\newblock In \emph{Proceedings of the IEEE/CVF International Conference on
  Computer Vision (ICCV)}, pages 10895--10904, 2023.

\bibitem[Luo et~al.(2025)Luo, Yuan, Wang, Li, Chen, Casta{\~n}eda, Cao, Li,
  Minor, Ben, Da, Ding, Hogg, Song, Lim, Jeong, He, Xue, Xiao, Wang, Yuen,
  Kautz, Chang, Iqbal, Fan, and Zhu]{SONIC}
Zhengyi Luo, Ye~Yuan, Tingwu Wang, Chenran Li, Sirui Chen, Fernando
  Casta{\~n}eda, Zi-Ang Cao, Jiefeng Li, David Minor, Qingwei Ben, Xingye Da,
  Runyu Ding, Cyrus Hogg, Lina Song, Edy Lim, Eugene Jeong, Tairan He, Haoru
  Xue, Wenli Xiao, Zi~Wang, Simon Yuen, Jan Kautz, Yan Chang, Umar Iqbal, Linxi
  Fan, and Yuke Zhu.
\newblock {SONIC}: Supersizing motion tracking for natural humanoid whole-body
  control.
\newblock \emph{arXiv preprint arXiv:2511.07820}, 2025.

\bibitem[Mahmood et~al.(2019)Mahmood, Ghorbani, Troje, Pons-Moll, and
  Black]{AMASS:ICCV:2019}
Naureen Mahmood, Nima Ghorbani, Nikolaus~F. Troje, Gerard Pons-Moll, and
  Michael~J. Black.
\newblock {AMASS}: Archive of motion capture as surface shapes.
\newblock In \emph{International Conference on Computer Vision}, pages
  5442--5451, October 2019.

\bibitem[Mannor et~al.(2003)Mannor, Rubinstein, and Gat]{mannor2003cross}
Shie Mannor, Reuven~Y Rubinstein, and Yohai Gat.
\newblock The cross entropy method for fast policy search.
\newblock In \emph{Proceedings of the 20th international conference on machine
  learning (ICML-03)}, pages 512--519, 2003.

\bibitem[Mu et~al.(2025)Mu, Ling, Shi, Ojeda, Xi, Shu, Zinno, and
  Peng]{StableMotion}
Yuxuan Mu, Hung~Yu Ling, Yi~Shi, Ismael~Baira Ojeda, Pengcheng Xi, Chang Shu,
  Fabio Zinno, and Xue~Bin Peng.
\newblock {StableMotion}: Training motion cleanup models with unpaired
  corrupted data.
\newblock In \emph{SIGGRAPH Asia 2025 Conference Papers}, 2025.
\newblock \doi{10.1145/3757377.3763817}.

\bibitem[Pan et~al.(2025)Pan, Wang, Qi, Liu, Bharadhwaj, Sharma, Wu, Shi,
  Malik, and Hogan]{SPIDER}
Chaoyi Pan, Changhao Wang, Haozhi Qi, Zixi Liu, Homanga Bharadhwaj, Akash
  Sharma, Tingfan Wu, Guanya Shi, Jitendra Malik, and Francois Hogan.
\newblock {SPIDER}: Scalable physics-informed dexterous retargeting.
\newblock \emph{arXiv preprint arXiv:2511.09484}, 2025.

\bibitem[Peng et~al.(2018)Peng, Abbeel, Levine, and van~de Panne]{DeepMimic}
Xue~Bin Peng, Pieter Abbeel, Sergey Levine, and Michiel van~de Panne.
\newblock {DeepMimic}: Example-guided deep reinforcement learning of
  physics-based character skills.
\newblock \emph{ACM Transactions on Graphics}, 37\penalty0 (4), 2018.
\newblock \doi{10.1145/3197517.3201311}.

\bibitem[Peng et~al.(2021)Peng, Ma, Abbeel, Levine, and Kanazawa]{AMP}
Xue~Bin Peng, Ze~Ma, Pieter Abbeel, Sergey Levine, and Angjoo Kanazawa.
\newblock {AMP}: Adversarial motion priors for stylized physics-based character
  control.
\newblock \emph{ACM Transactions on Graphics}, 2021.

\bibitem[Qi et~al.(2026)Qi, Chen, Liu, Lin, Lian, Liang, Zhang, Guan, Wang,
  Zhang, Yu, Wang, and Yi]{humanoid-gpt26}
Zekun Qi, Xuchuan Chen, Dairu Liu, Chenghuai Lin, Yunrui Lian, Sikai Liang,
  Zhikai Zhang, Yu~Guan, Jilong Wang, Wenyao Zhang, Xinqiang Yu, He~Wang, and
  Li~Yi.
\newblock Humanoid-{GPT}: Scaling data and structure for zero-shot motion
  tracking.
\newblock \emph{arXiv preprint arXiv:2606.03985}, 2026.

\bibitem[Rempe et~al.(2021)Rempe, Birdal, Hertzmann, Yang, Sridhar, and
  Guibas]{HuMoR}
Davis Rempe, Tolga Birdal, Aaron Hertzmann, Jimei Yang, Srinath Sridhar, and
  Leonidas~J. Guibas.
\newblock {HuMoR}: {3D} human motion model for robust pose estimation.
\newblock In \emph{ICCV}, 2021.

\bibitem[Shin et~al.(2024)Shin, Kim, Halilaj, and Black]{shin2024wham}
Soyong Shin, Juyong Kim, Eni Halilaj, and Michael~J. Black.
\newblock {WHAM}: Reconstructing world-grounded humans with accurate {3D}
  motion.
\newblock In \emph{IEEE/CVF Conference on Computer Vision and Pattern
  Recognition (CVPR)}, pages 2070--2080, 2024.

\bibitem[Tessler et~al.(2024)Tessler, Guo, Nabati, Chechik, and
  Peng]{MaskedMimic}
Chen Tessler, Yunrong Guo, Ofir Nabati, Gal Chechik, and Xue~Bin Peng.
\newblock {MaskedMimic}: Unified physics-based character control through masked
  motion inpainting.
\newblock \emph{ACM Transactions on Graphics}, 2024.

\bibitem[Wang et~al.(2026{\natexlab{a}})Wang, Liao, Zhang, Ren, Sreenath, and
  Xiong]{SPARK}
Hanwen Wang, Qiayuan Liao, Bike Zhang, Kunzhao Ren, Koushil Sreenath, and
  Xiaobin Xiong.
\newblock {SPARK}: Skeleton-parameter aligned retargeting on humanoid robots
  with kinodynamic trajectory optimization.
\newblock \emph{arXiv preprint arXiv:2603.11480}, 2026{\natexlab{a}}.

\bibitem[Wang et~al.(2026{\natexlab{b}})Wang, Zhu, Zhi, Li, Li, Li, Xiao, Wang,
  Jia, and Huang]{wang2026omnixtreme}
Yunshen Wang, Shaohang Zhu, Peiyuan Zhi, Yuhan Li, Jiaxin Li, Yong-Lu Li,
  Yuchen Xiao, Xingxing Wang, Baoxiong Jia, and Siyuan Huang.
\newblock {OmniXtreme}: Breaking the generality barrier in high-dynamic
  humanoid control.
\newblock \emph{arXiv preprint arXiv:2602.23843}, 2026{\natexlab{b}}.

\bibitem[Xue et~al.(2025)Xue, Pan, Yi, Qu, and Shi]{DIALMPC}
Haoru Xue, Chaoyi Pan, Zeji Yi, Guannan Qu, and Guanya Shi.
\newblock Full-order sampling-based {MPC} for torque-level locomotion control
  via diffusion-style annealing.
\newblock In \emph{IEEE International Conference on Robotics and Automation},
  pages 4974--4981, 2025.
\newblock \doi{10.1109/ICRA55743.2025.11127320}.

\bibitem[Yang et~al.(2026)Yang, Huang, Wu, Kanazawa, Abbeel, Sferrazza, Liu,
  Duan, and Shi]{OmniRetarget}
Lujie Yang, Xiaoyu Huang, Zhen Wu, Angjoo Kanazawa, Pieter Abbeel, Carmelo
  Sferrazza, C.~Karen Liu, Rocky Duan, and Guanya Shi.
\newblock {OmniRetarget}: Interaction-preserving data generation for humanoid
  whole-body loco-manipulation and scene interaction.
\newblock In \emph{IEEE International Conference on Robotics and Automation},
  2026.
\newblock \doi{10.1109/ICRA57385.2026.11696487}.

\bibitem[Yuan et~al.(2023)Yuan, Song, Iqbal, Vahdat, and Kautz]{PhysDiff}
Ye~Yuan, Jiaming Song, Umar Iqbal, Arash Vahdat, and Jan Kautz.
\newblock {PhysDiff}: Physics-guided human motion diffusion model.
\newblock In \emph{ICCV}, 2023.

\bibitem[Ze et~al.(2025)Ze, Zhao, Wang, Kanazawa, Duan, Abbeel, Shi, Wu, and
  Liu]{ze2025twist2}
Yanjie Ze, Siheng Zhao, Weizhuo Wang, Angjoo Kanazawa, Rocky Duan, Pieter
  Abbeel, Guanya Shi, Jiajun Wu, and C.~Karen Liu.
\newblock {TWIST2}: Scalable, portable, and holistic humanoid data collection
  system.
\newblock \emph{arXiv preprint arXiv:2511.02832}, 2025.

\bibitem[Zhang et~al.(2024)Zhang, Bhatnagar, Xu, Winkler, Kadlecek, Tang, and
  Bogo]{RoHM}
Siwei Zhang, Bharat~Lal Bhatnagar, Yuanlu Xu, Alexander Winkler, Petr Kadlecek,
  Siyu Tang, and Federica Bogo.
\newblock {RoHM}: Robust human motion reconstruction via diffusion.
\newblock In \emph{Proceedings of the IEEE/CVF Conference on Computer Vision
  and Pattern Recognition (CVPR)}, pages 14606--14617, 2024.

\bibitem[Zhao et~al.(2026)Zhao, Li, Yang, Wang, Zhao, Lu, Zhang, Shen, Long,
  and Cao]{zhao2026maketrackingeasy}
Qingrui Zhao, Yiheng Li, Kaiyue Yang, Xiyu Wang, Shiqi Zhao, Yi~Lu, Xinfang
  Zhang, Qiu Shen, Xiao-Xiao Long, and Xun Cao.
\newblock Make tracking easy: Neural motion retargeting for humanoid whole-body
  control.
\newblock \emph{arXiv preprint arXiv:2603.22201}, 2026.

\end{thebibliography}

\end{document}